\documentclass[12pt,a4paper]{article} 
\usepackage[utf8]{inputenc}
\usepackage{graphicx}
\usepackage{caption}
\usepackage{subcaption}
\usepackage{nameref,hyperref}

\usepackage[right]{lineno}

\begin{document}

\begin{flushleft}
{\Large
\textbf{\newline{Deep Learning for Audio Transcription on Low-Resource Datasets}}
}
\newline
\\
\textbf{Veronica Morfi*, Dan Stowell}
\\
\bigskip
\textbf{Machine Listening Lab, Centre for Digital Music (C4DM), Queen Mary University of London, UK}
\\
\bigskip
\textbf{*g.v.morfi@qmul.ac.uk}

\end{flushleft}



\abstract{In training a deep learning system to perform audio transcription, two practical problems may arise. Firstly, most datasets are \textit{weakly} labelled, having only a list of events present in each recording without any temporal information for training. Secondly, deep neural networks need a very large amount of labelled training data to achieve good quality performance, yet in practice it is difficult to collect enough samples for most classes of interest. In this paper, we propose factorising the final task  of audio transcription into multiple intermediate tasks in order to improve the training performance when dealing with this kind of low-resource datasets. We evaluate three data-efficient approaches of training a stacked convolutional and recurrent neural network for the intermediate tasks. Our results show that different methods of training have different advantages and disadvantages.}






\section{Introduction}
\label{sec:intro}
Machine learning has experienced a strong growth in recent years, due to increased dataset sizes and computational power, and to advances in deep learning methods that can learn to make predictions in extremely nonlinear problem settings \cite{LeCun:15}. However, a large amount of data is needed in order to train a neural network that can achieve a good quality performance. With the increased amount of audio datasets publicly available there is also an increase of tagging labels available for them. We refer to these tagging labels, that only indicate the presence or not of a type of event in a recording and lack any temporal information about it, as \textit{weak} labels. 

A lot of research has been done in tagging of audio recordings. In \cite{Choi:16}, the authors proposed a content-based automatic music tagging algorithm
using deep convolutional neural networks. In \cite{Xu:17b}, the authors proposed to use a shrinking deep neural network incorporating unsupervised feature learning to handle the multi-label audio tagging. Furthermore, considering that only chunk level rather than frame-level labels are available, a large set of contextual frames of the chunk were fed into the network to perform this task. In \cite{Xu:17a, Adavanne:17a}, the authors use a stacked convolutionla recurrent network to perform environmental audio tagging and tag the presence of birdsong, respectively. While in \cite{Pons:17}, the authors explore two different models for end-to-end music audio tagging when there is a large amount of training data. 

However, in recent decades, there has also been an increase to the demand of transcription predictions for a variety of audio recordings instead of just the tags of a recording. Some potential applications where audio event transcription is necessary are context awareness for cars, mobiles, etc., surveillance for dangerous events and crimes, analysis and monitoring of biodiversity, recognition of noise sources and machine faults and many more. Depending on the audio event to be detected and classified in each task it may become difficult to collect enough samples for them. Furthermore, different tasks use task specific datasets, hence the amount of recordings available may be limited. Additionally, annotating data with \textit{strong} labels, labels that contain temporal information about the events, to train transcription predictors is a time consuming process involving a lot of manual labour. On the other hand, collecting weakly labelled data takes much less time, since the annotator only has to mark the active sound event classes and not their exact boundaries. We refer to datasets that only have this type of weak labels, may contain rare events and have limited amounts of training data as \textit{low-resource} datasets.

In comparison to supervised techniques that are trained on strong labels, there has been relatively little work on learning to perform audio event transcription using weakly labelled data. In \cite{Briggs:12, Ruiz:15} the authors try to exploit weak labels in birdsong detection and bird species classification, while in \cite{Fanioudakis:17} the authors use deep networks to tag the location of bird vocalisations. In \cite{Schluter:16} singing voice is pinpointed from weakly labelled examples. In \cite{Kong:17}, the authors used a joint detection-classification network that slices the audio into blocks and a audio detector and classification on each block then uses the overall audio tag to train using the weak labels of a recording. Furthermore, in \cite{Adavanne:17b} the authors train a network that can do automatic scene transcription from weak labels and in \cite{Hershey:17} audio from YouTube videos is used in order to train and compare different previously proposed convolutional neural network architectures for audio event detection and classification. Finally, in \cite{Kumar:16, Kumar:17} the authors use weakly labelled data for audio event detection in order to move from the weak labels space to strong labels. Most of these methods formulate the provided weak labels of the recordings into a multi instance learning (MIL) problem. However, for the methods using neural networks, none of the datasets used could be considered low-resource. Most of the datasets used either come from transcription/detection challenges (e.g. DCASE) or online sources, such as Youtube or xeno-canto, that contain a large number of training data. 

Training a neural network to predict an audio transcription using a low-resource dataset can sometimes prove to be impossible. A network needs to have enough parameters to be able to predict all the different classes without ignoring any rare events, but also be small enough or have just the right amount of regularisation as to not overfit the limited amount of training data available. This becomes even harder when the task is a weak-to-strong prediction where the network needs to predict full transcriptions from weak labels. Unfortunately, there is no specific way of defining a network and type of training that ensures that a transcription will be predicted successfully. However, a full transcription task can be defined as multiple intermediate tasks of detection and classification that might be easier to train even when using a low-resource dataset. 

In this paper, we propose a factorisation of the final full transcription task into multiple simpler intermediate tasks of audio event detection and audio tagging in order to predict an intermediate transcription that can be used to boost the performance of the full transcription task. For each intermediate task we propose a training setup to optimise their performance. Finally, we train the intermediate tasks independently and in two multi-task learning settings and compare their results.

The rest of the paper is structured as follows: Section \ref{sec:fact} describes the way we factorise the transcription task into intermediate tasks and  presents in detail our setup and network architectures. In Section \ref{sec:approaches} we propose three different training approaches for the intermediate tasks, two of which are implemented in a multi-task learning setting. In Section \ref{sec:eval}, we present our experiments and compare the results of each training approach. Finally, in Section \ref{sec:discussion}, we discuss our findings and future research directions.

\section{Task Factorisation}
\label{sec:fact}
\begin{figure}[!htb]
\centering
\includegraphics[width=15cm]{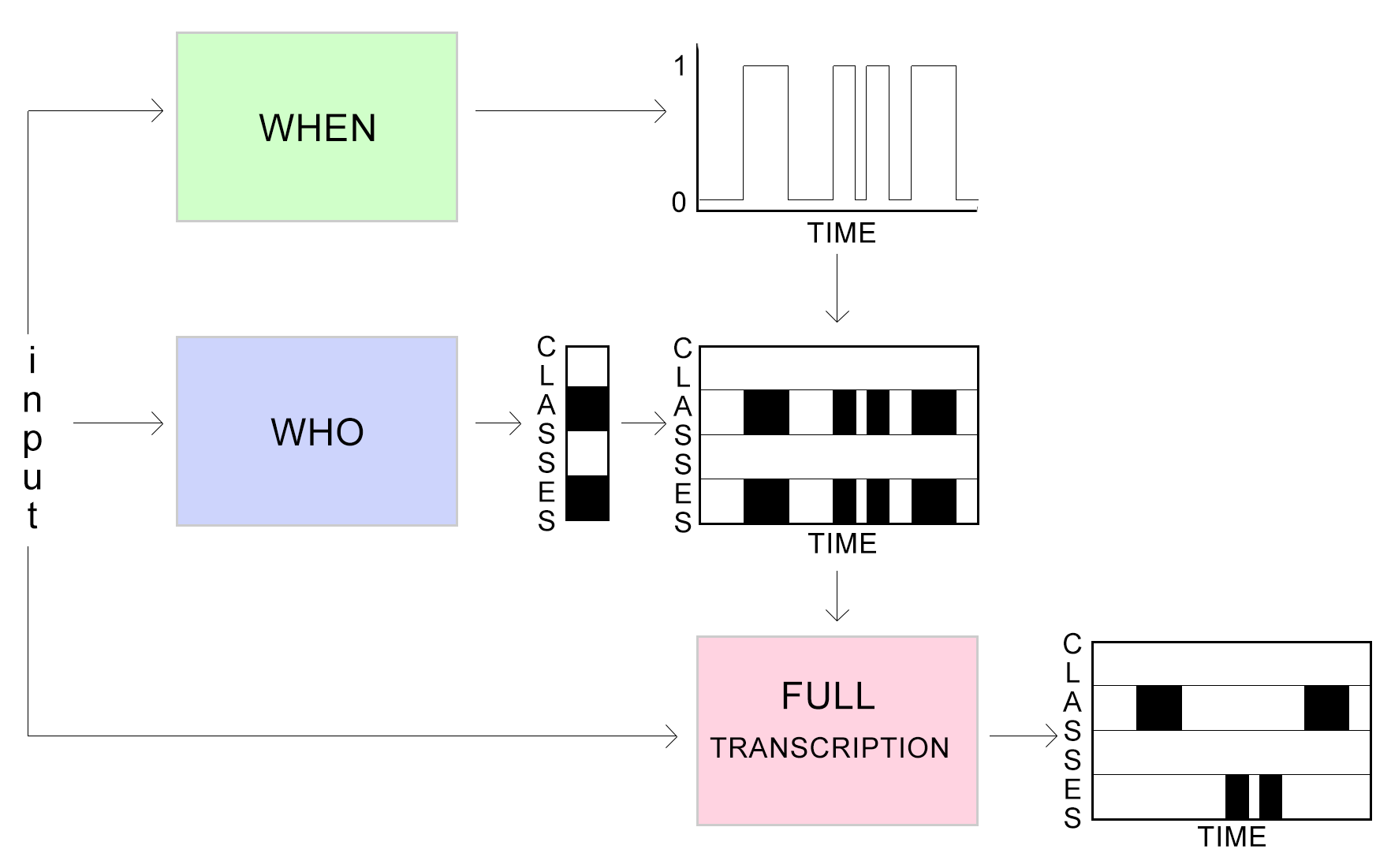}
\caption{Factorisation of the full transcription task. WHEN network performs audio event detection considering all labels as one label. WHO network performs audio tagging for all available labels. The predictions of WHEN and WHO produce an intermediate transcription that is used to boost the performance of the full transcription network.}
\label{fig:bigidea}
\end{figure}

A full audio transcription task can be described as audio event detection followed by event classification. In order to properly train a full transcription network we need a large amount of data which is not available in a low-resource dataset. Since it is very hard to train a network to predict full transcription on a low-resource dataset, we factorise the final task of full transcription into intermediate tasks that can predict an intermediate transcription matrix that can later be used to boost the performance of a full transcription network. Figure \ref{fig:bigidea} depicts the overall task factorisation into the intermediate tasks and how they interact with the final task of full transcription. We define a WHEN network that performs audio event detection considering all classes as one general class and predicts when any event is present without taking into consideration the different event classes. We also define a WHO network that performs audio tagging without predicting any temporal information. By combining the two different predictions from these networks we create an intermediate transcription that provides us with the events present in a recording and the times where any of these events could be present in a recording. This intermediate transcription is to be used as supplementary information when training the full transcription network in order to improve its performance by focusing its attention to the classes present in a recording and the time frames that may contain them.

When using a large enough dataset that provides satisfactory training data and has a a good representation for each different class, many methods have been successful in performing both of the intermediate tasks. A few methods for audio event detection can be found in \cite{Fanioudakis:17} and \cite{Schluter:16}, while for audio tagging in \cite{Kong:17, Xu:17b, Xu:17a, Adavanne:17a, Pons:17, Choi:16}. These tasks are less challenging to train for than a full transcription task. However, using a low-resource dataset can degrade their performance. Hence, in order to achieve a satisfactory performance when training with a low-resource dataset, we propose a few training setups and techniques. The rest of this chapter describes in detail the task specific setups and techniques that we used.

\subsection{Input Features}
As input to all our intermediate networks, log mel-band energy is extracted from audio in 23ms Hamming windows with 50\% overlap. In order to do so the \texttt{librosa} Python library is used. In total, 40 mel-bands are used in the 0--44100 Hz range. For a given 5 second audio input, the feature extraction produces a $T$x40 output ($T = 432$).

\subsection{Audio Event Detection (WHEN)}
\label{ssec:when}
In our proposed task factorisation, the WHEN network performs single class audio event detection as the first intermediate task towards full transcription. For a multi-class dataset, one would have to train a separate network for each class in order to perform single class event detection. However, in a low-resource dataset, training an audio event detector for each class can be nearly impossible. The number of classes might be too large, making it a time consuming task. Furthermore, some of the classes might have very rare occurrences, limited to only a couple of recordings, hence making it infeasible to train a neural network for them. Nevertheless, many low-resource datasets are usually used for discriminating subclasses of a general class e.g. song of different bird species, sound of different car engines, barking of different dog breeds, notes produced by an instrument. These subclasses usually share some common features and characteristics, hence in order to achieve a good performance in the audio event detection task, we propose to consider all subclasses as one general class and train a single WHEN network to perform single class event detection. This reduces the training time compared to training one network for each subclass and also solves any training issues caused by rare events. 

\subsubsection{Neural Network Architecture}
\label{sssec:whenarch}
For our audio event detector we use a state-of-the-art stacked convolutional and recurrent neural network architecture. Table \ref{tab:when} describes the parameters of the proposed architecture. 

\begin{table}[!htb]
\caption{WHEN network architecture. Size refers to either kernel shape or number of units. \#Fmaps is the number of feature maps in the layer. Activation denotes the activation used for the layer and l2\_regularisation the amount of l2 kernel regularisation used in the layer.}
\centering
\begin{tabular}{lcccc}
\hline
\textbf{Layer}	& \textbf{Size}	& \textbf{\#Fmaps} & \textbf{Activation} & \textbf{l2\_regularisation}\\
\hline
Convolution 2D		& 3x3			& 64 & Linear & 0.001\\
Batch Normalisation		& -			& - & - & -\\
Activation & - & - & ReLU & -\\ 
Convolution 2D		& 3x3			& 64 & Linear & 0.001\\
Batch Normalisation		& -			& - & - & -\\
Activation & - & - & ReLU & -\\ 
Max Pooling & 1x5 & - & - & -\\
Convolution 2D		& 3x3			& 64 & Linear & 0.001\\
Batch Normalisation		& -			& - & - & -\\
Activation & - & - & ReLU & -\\ 
Convolution 2D		& 3x3			& 64 & Linear & 0.001\\
Batch Normalisation		& -			& - & - & -\\
Activation & - & - & ReLU & -\\ 
Max Pooling & 1x4 & - & - & - \\
Convolution 2D		& 3x3	& 64 & Linear & 0.001\\
Batch Normalisation		& -			& - & - & -\\
Activation & - & - & ReLU & -\\ 
Convolution 2D		& 3x3			& 64 & Linear & 0.001\\
Batch Normalisation		& -	& - & - &-\\
Activation & - & - & ReLU & -\\ 
Max Pooling & 1x2 & - & - & -\\
Reshape & - & - & - & -\\
Bidirectional GRU & 64 & - & tanh & 0.01\\
Bidirectional GRU & 64 & - & tanh & 0.01\\
Time Distributed Dense & 64 & - & ReLU & 0.01\\
Time Distributed Dense & 1 & - & Sigmoid & 0.01\\
Flatten & - & - & - & -\\ 
\hline
\end{tabular}
\label{tab:when}
\end{table}

The log mel-band energy feature extracted from the audio is fed to the neural network, which sequentially produces the predicted strong labels for each recording. The input to the proposed network is a $T$x40 feature matrix. The convolutional layers in the beginning of the network are in charge of learning the local shift-invariant features of this input. We use a 3x3 receptive field and the padding arguments set as `same' in order to maintain the same size as the input in all our convolutional layers. The max-pooling operation is performed along the frequency axis after every convolutional layer to reduce the dimension for the feature matrix while preserving the number of frames $T$. The output of the convolutional part of the network is then fed to bi-directional gated recurrent units (GRUs) with tanh activation to learn the temporal structure of audio events. Next we apply time distributed dense layers to reduce feature-length dimensionality. Note that the time resolution of $T$ frames is maintained in both the GRU and dense layers. A sigmoid activation is used in the last time-distributed dense layer to produce a binary prediction of whether there is an event present in each time frame. This prediction layer outputs the strong labels for a recording. The dimensions of each prediction are $T$x1. Finally, we calculate the loss on this output as explained in Section \ref{ssec:loss}.

\subsubsection{Multi Instance Learning}
\label{ssec:loss}
When used for training audio event detectors, low-resource datasets present the issue of weak-to-strong prediction. Low-resource datasets only provide the user with weak labels, labels that don't include any temporal information about the events but only denote the presence or absence of a specific class in a recording. However, audio event detectors produce instance labels referred to as strong labels, hence provide full temporal information about the events in a recording. 

The most common way to train a network for weak-to-strong prediction is the multi instance learning (MIL) setting. The concept of MIL was first properly developed in \cite{Dietterich:97} for drug activity detection. MIL is described in terms of \textit{bags}, with a bag being a collection of instances. The existing weak labels are attached to the bags, rather than the individual instances within them. Positive bags have at least one positive instance, an instance for which the target class is active. On the other hand, negative bags contain negative instances only, the target class is not active in them. A negative bag is thus pure while a positive bag is presumably impure, since the latter most likely contains both positive and negative instances. Hence, all instances in a negative bag can be uniquely assigned a negative label but for a positive bag this cannot be done. There is no direct knowledge of whether an instance in a positive bag is positive or negative. Thus, it is the bag-label pairs and not the instance-label pairs which form the training data, and from which a classifier which classifies individual instances must be learned.

Let the training data be composed of $N$ bags, i.e. $\left\lbrace B_1, B_2, ..., B_N \right\rbrace$, the $i$-th bag is composed of $M_i$ instances, i.e. $\left\lbrace B_{i1}, B_{i2}, ..., B_{iM_i} \right\rbrace$, where each instance is a $p$-dimensional feature vector, e.g. the $j$-th instance of the $i$-th bag is $\left[ B_{ij1}, B_{ij2}, ..., B_{ijp} \right] ^T$. We represent the bag-label pairs as $\left( B_i, Y_i \right)$, where $Y_i \in \left\lbrace 0, 1 \right\rbrace$ is the bag label for bag $B_i$. $Y_i = 0$ denotes a negative bag and $Y_i = 1$ denotes a positive bag. 

One na\"ive but commonly used way of inferring the individual instances' labels from the bag labels is assigning the bag label to each instance of that bag: we refer to this method as \textit{false strong labelling}. During training, a neural network in the MIL setting with false strong labels tries to minimise the average divergence between the network output for each instance and the false strong labels assigned to them, identically to an ordinary supervised learning scenario. However, it is evident that the false strong labelling approach is an approximation of the loss for a strong label prediction task, hence it has some disadvantages. When using false strong labels some kind of early stopping is necessary since when perfect accuracy is achieved that would mean all positive instance predictions for a positive bag. However, there is no clear way of defining a specific point for early stopping. This is the same issue that all methods in the MIL setting face. As mentioned before a positive bag might include both positive and negative instances, however false strong labels will force the network towards positive predictions for both. Additionally, by using strong false labels there is an imbalance of positive and negative instance labels compared to the true strong labels, since a substantial amount of negative instances are considered as positive during training. Finally, a negative instance may appear in both a negative and positive recording, however due to the false labelling of negative instances as positive in positive bags, the network may not learn the proper prediction for this kind of instance.

As an alternative to false strong labels, one can attempt to infer labels of individual instances in bag $B_i$ by making a few educated assumptions. The most common ones are: if $Y_i = 0 $, all instances of bag $B_i$ are negative instances, hence $y_{ij} = 0, \forall j$, while on the other hand, if $Y_i = 1$, at least one instance of bag $B_i$ is equal to one. For all instances of bag $B_i$, this relation between the bag label and instance labels can be simply written as:

\begin{equation}
Y_i = \max_j y_{ij}
\label{eq:milmax}
\end{equation}

The conventional way of training a neural network for strong labelling is providing instance specific (strong) labels for a collection of training instances. Training is performed by updating the network weights to minimize the average divergence between the network output in response to these instances and the desired output, the ground truth of the training instances. In the MIL setting using equation (\ref{eq:milmax}) to define a characteristic of the strong labels, we must modify the manner in which the divergence to be minimized is computed, to utilize only weak labels, as proposed in \cite{Zhou:02}. 

Let $o_{ij}$ represent the output of the network for input $B_{ij}$, the $j$-th instance in $B_i$, the $i$-th bag of training instances. We define the bag-level divergence for bag $B_i$ as:

\begin{equation}
E_i = \frac{1}{2} \left( \max_{1 \leq j \leq M_j} (o_{ij}) - Y_i \right) ^2
\label{eq:E_i}
\end{equation}

\noindent where $Y_i$ is the label assigned to bag $B_i$. 

The overall divergence on the training set is obtained by summing the divergences of all the bags in the set:

\begin{equation}
E = \sum_{i=1}^N E_i
\label{eq:E}
\end{equation}

Equation (\ref{eq:E_i}) indicates that if at least one instance of a positive bag is perfectly predicted as positive, or all the instances of a negative bag are perfectly predicted as negative, then the error on the concerned bag is zero. Otherwise, the weights will be updated according to the error on the instance whose corresponding actual output is the maximal among all the instances in the bag. Note that such an instance is typically the most easy to be predicted as positive for a positive bag, while it is the most difficult to be predicted as negative for a negative bag. It seems that this sets a low burden on producing a positive output but a strong burden on producing a negative output. As indicated in \cite{Amar:01}, the value of a bag is fully determined by its instance with the maximal output, regardless how many real positive or negative instances in the bag. Therefore, in fact the burden on producing a positive or negative output is not unbalanced, at least at bag-level. However, on an instance-level, when using max to compute the loss, only one instance per bag contributes to the gradient, which may lead to inefficient training. Additionally, as mentioned earlier, in positive bags the network only has to accurately predict the label for the easiest positive instance to reach a perfect accuracy, thus not paying as much attention to the rest of the positive instances that might be harder to accurately detect.

In order to train our proposed WHEN network, we want all predictions to weigh in on the loss and not just the one with the maximum value, as is the case with MIL using max. In \cite{Liu:17}, the authors proposed the ``noisy-or'' pooling function to be used instead of max. However, noisy-or has been proven to not perform as well as max for audio event detection \cite{Wang:18}. As discussed in \cite{Wang:18}, a significant problem with noisy-or is that the label of a bag is computed via the product of the instance labels as seen in equation (\ref{eq:noisy-or}). This calculation relies heavily on the assumed conditional independence of instance labels, an assumption which is highly inaccurate in audio event detection. Furthermore, this can lead the system to believe a bag is positive even though all its instances are negative. 

\begin{equation}
Y_i = 1 - \prod_{1 \leq j \leq M_j}(1-y_{ij})
\label{eq:noisy-or}
\end{equation}

Using all instances in a bag for computation of the loss and backpropagated gradient is important, since the network ideally should acquire some knowledge from every instance in every epoch. However, it is hard to find an elegant theoretical interpretation of the characteristics of the instances in a bag. On the other hand, we propose a couple simpler assumptions about these characteristics that can achieve a similar effect. One assumption is to consider the mean of the instance predictions of a bag. If a bag is negative the mean should be zero, while if it is positive it should be greater than zero. The true mean is unknown in weakly labelled data. A na\"ive assumption is to presume that approximately half of the time a specific event will be present in a recording. Even though this is not true all of the time, it takes into consideration the predictions for all instances, and also inserts a bias to the loss that will keep producing gradient for training even after the max term has reached its perfect accuracy. However, this is indeed a na\"ive assumption that will guide the network to predict a balanced amount of positives and negatives which may make it more sensitive to all kind of audio events, even when they are not the ones in question.

Another simple yet accurate assumption is that on both negative and positive recordings the minimum predictions at an instance-level should be zero. It is possible for a positive recording to have no negative frames however it is extremely rare in practice. This assumption could be used in synergy with max and mean to enforce the prediction of negative instances even on positive recordings and manage a certain level of the bias that is introduced with considering mean in the computation of the loss. 

We train a network on a loss function that takes into account all the above mentioned assumptions and compute the max, mean and min from the predictions of a recording and depending on whether a recording is positive or negative we predict their divergence from different conditions. 

Our proposed loss function is computed as:

\begin{equation}
Loss = \frac{1}{3} \big( bin\_cr(max_j (o_{ij}), Y_i) + bin\_cr(mean_j (o_{ij}), \frac{Y_i}{2}) + bin\_cr(min_j (o_{ij}), 0) \big)
\label{eq:mmmloss}
\end{equation}

\noindent where $bin\_cr(x,y)$ is a function that computes the binary cross-entropy between $x$ and $y$, $o_{ij}$ are all the predicted strong labels of bag $B_i$, where $j = 1...M_i$ with $M_i$ being the total number of instances in a bag, and $Y_i$ is the label of the bag. 

\begin{figure}
\centering
\begin{subfigure}[!htb]{0.4\textwidth}
   \includegraphics[width=1\linewidth]{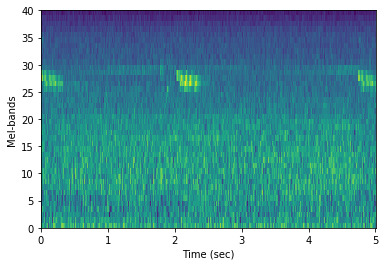}
\end{subfigure}
\begin{subfigure}[!htb]{0.4\textwidth}
   \includegraphics[width=1\linewidth]{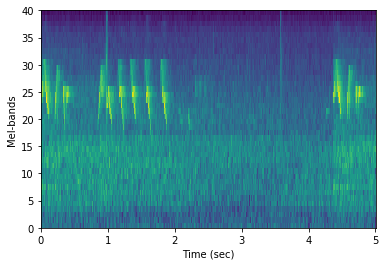}
\end{subfigure}\\
\begin{subfigure}[!htb]{0.4\textwidth}
   \includegraphics[width=1\linewidth]{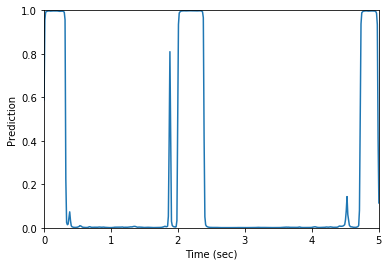}
   \caption{MIL using max}
   \label{623max}
\end{subfigure}
\begin{subfigure}[!htb]{0.4\textwidth}
   \includegraphics[width=1\linewidth]{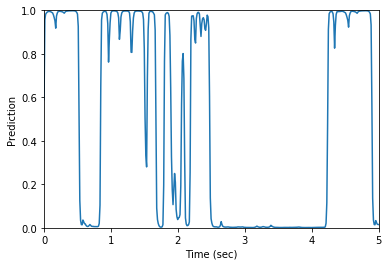}
   \caption{MIL using max}
      \label{667max}
\end{subfigure}\\
\begin{subfigure}[!htb]{0.4\textwidth}
   \includegraphics[width=1\linewidth]{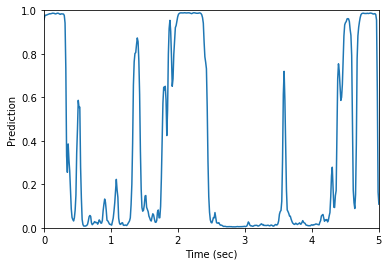}
   \caption{MIL using MMM}
   \label{623mmm}
\end{subfigure}
\begin{subfigure}[!htb]{0.4\textwidth}
   \includegraphics[width=1\linewidth]{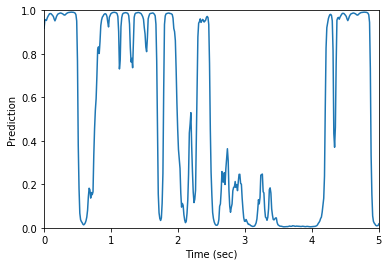}
   \caption{MIL using MMM}
   \label{667mmm}
\end{subfigure}
\caption{Predicted transcription, of two recordings. Figures \ref{623max} and \ref{667max} depict the results of our WHEN network trained with max loss. Figures \ref{623mmm} and \ref{667mmm} depict the results of our WHEN network trained with MMM loss.}
\label{fig:maxvsmmm}
\end{figure}

We refer to this as an \textit{MIL setting using MMM}. For negative recordings, equation (\ref{eq:mmmloss}) will compute the binary cross-entropy between the max, mean and min of the predictions of the instances of a bag $B_i$ and zero. This denotes that the predictions for all instances of a negative recording should be zero. On the other hand, for positive recordings the predictions should span the full dynamic range from zero to one, biased towards a similar amount of positive and negative instances. Our proposed loss function is designed to balance the positive and negative predictions in a bag resulting in a network that has the flexibility of learning from harder-to-predict positive instances even after many epochs. This is due to the fact that there are no obvious local minima to get stuck in as in the max case. Some examples of the difference between the predictions produced by MIL using max and MIL using MMM when our proposed WHEN network is trained for birdsong detection, are depicted in Figure \ref{fig:maxvsmmm}. It becomes apparent that MIL using MMM can correctly classify harder to predict instances, especially when studying the difference between Figures \ref{623max} and \ref{623mmm}. In Figure \ref{623mmm}, one can notice that the network is able to correctly classify the harder to predict instances between the three main audio events. 

\subsubsection{Half and Half training}
In the MIL setting for weak-to-strong labelling, it is of great importance to have a good balance between positive and negative bags, in order for the network to be able to distinguish what can be considered a positive instance and what a negative one. A simple approach to achieve this kind of balanced training is to have balanced minibatches. In our approach, we implement this by duplicating negative or positive recordings randomly during training depending which ones are less in the whole dataset. Thus each minibatch during training will consist of the same amount of positive and negative recordings, which in our case is 4 positive and 4 negative recordings. We call this kind of input Half and Half (HnH).

\subsection{Audio Tagging (WHO)}
\label{ssec:who}
The second intermediate task of our approach is the WHO network that performs audio tagging using the provided weak labels of a low-resource dataset. This task follows supervised training since the weak labels provided are the ones that the network will try to learn how to predict. Hence, there are no particular training techniques that we use for the WHO network. 

\subsubsection{Neural Network Architecture}
A similar network architecture to the one proposed for WHEN (see Table \ref{tab:when}) is used for the first few layers of WHO in order to implement our proposed training approaches that we introduce in Section \ref{sec:approaches}. Table \ref{tab:who} describes the structure of each individual layer used in the WHO network. 

\begin{table}[!htb]
\caption{WHO network architecture. Size refers to either kernel shape or number of units. \#Fmaps is the number of feature maps in the layer. Activation denotes the activation used for the layer and l2\_regularisation the amount of l2 kernel regularisation used in the layer.}
\centering
\begin{tabular}{lcccc}
\hline
\textbf{Layer}	& \textbf{Size}	& \textbf{\#Fmaps} & \textbf{Activation} & \textbf{l2\_regularisation}\\
\hline
Convolution 2D		& 3x3			& 64 & Linear & 0.001\\
Batch Normalisation		& -			& - & - & -\\
Activation & - & - & ReLU & -\\ 
Convolution 2D		& 3x3			& 64 & Linear & 0.001\\
Batch Normalisation		& -			& - & - & -\\
Activation & - & - & ReLU & -\\ 
Max Pooling & 1x5 & - & - & -\\
Convolution 2D		& 3x3			& 64 & Linear & 0.001\\
Batch Normalisation		& -			& - & - & -\\
Activation & - & - & ReLU & -\\ 
Convolution 2D		& 3x3			& 64 & Linear & 0.001\\
Batch Normalisation		& -			& - & - & -\\
Activation & - & - & ReLU & -\\ 
Max Pooling & 1x4 & - & - & - \\
Convolution 2D		& 3x3	& 64 & Linear & 0.001\\
Batch Normalisation		& -			& - & - & -\\
Activation & - & - & ReLU & -\\ 
Convolution 2D		& 3x3			& 64 & Linear & 0.001\\
Batch Normalisation		& -	& - & - &-\\
Activation & - & - & ReLU & -\\ 
Max Pooling & 1x2 & - & - & -\\
Global Average Pooling 2D & - & - & - & -\\
Dense & \#labels & - & Sigmoid & 0.001\\
\hline
\end{tabular}
\label{tab:who}
\end{table}

Similar to the WHEN network, the log mel-band energy feature extracted from the audio is used as input with shape $T$x40, where $T$ is the number of time frames in a recording. The convolutional layers in the beginning of the network are in charge of learning the local shift-invariant features of this input. We use a 3x3 receptive field and the padding arguments set as `same'. Max-pooling is performed along the frequency axis after every convolutional layer to reduce the dimension for the feature matrix. Global average pooling is finally applied to the output of the convolutional part of the network and the results are fed to a dense layer that has units equal to the number of labels for our tagging task with sigmoid activation that predict the probability of each class being present in a recording. The dimensions of each prediction are 1x\#labels. Finally, we calculate the binary cross-entropy loss on this output and the ground truth extracted from the weak labels. 

\section{Training Methods}
\label{sec:approaches}
Three different methods were used to train the two intermediate tasks. One of the them is the simple and usual approach of training each network independently for each task. Additionally, two multi-task learning (MTL) methods were tested, namely joint training and tied weights training, both of which follow a hard parameter sharing approach. All three different methods have advantages and disadvantages that will be compared in detail in Section \ref{sec:eval}.

MTL \cite{Caruana:97} aims to improve the performance of multiple learning tasks by sharing useful information among them. MTL can be very useful when using low-resource datasets since it can exploit useful information from other related learning tasks to help alleviate the issue of limited data. Based on the assumption that the multiple tasks are related, MTL is empirically and theoretically found to lead to better performance than independent learning. MTL is similar to transfer learning \cite{Pan:10} which also transfers knowledge from one task to another. However, the focus of transfer learning is to help a single target task by initially training on one or multiple tasks while MTL uses multiple tasks to help each other. Furthermore, MTL can be viewed as a generalization of multi-label learning \cite{Zhang:14} when different tasks in multi-task learning share the same training data. 
 
The motivation behind using MTL includes the implicit data augmentation, since a model that learns two tasks simultaneously is able to learn a more general representation. Also, if data is limited MTL can help the model focus its attention on those features that actually matter as other tasks will provide additional evidence for the relevance or irrelevance of those features. Finally, MTL acts as a regulariser by introducing an inductive bias that reduces the risk of overfitting. An overview of MTL can be found in \cite{Zhang:18}.

\subsection{Separate Training}
\label{ssec:separate}
\begin{figure}[!htb]
\centering
\includegraphics[width=8cm]{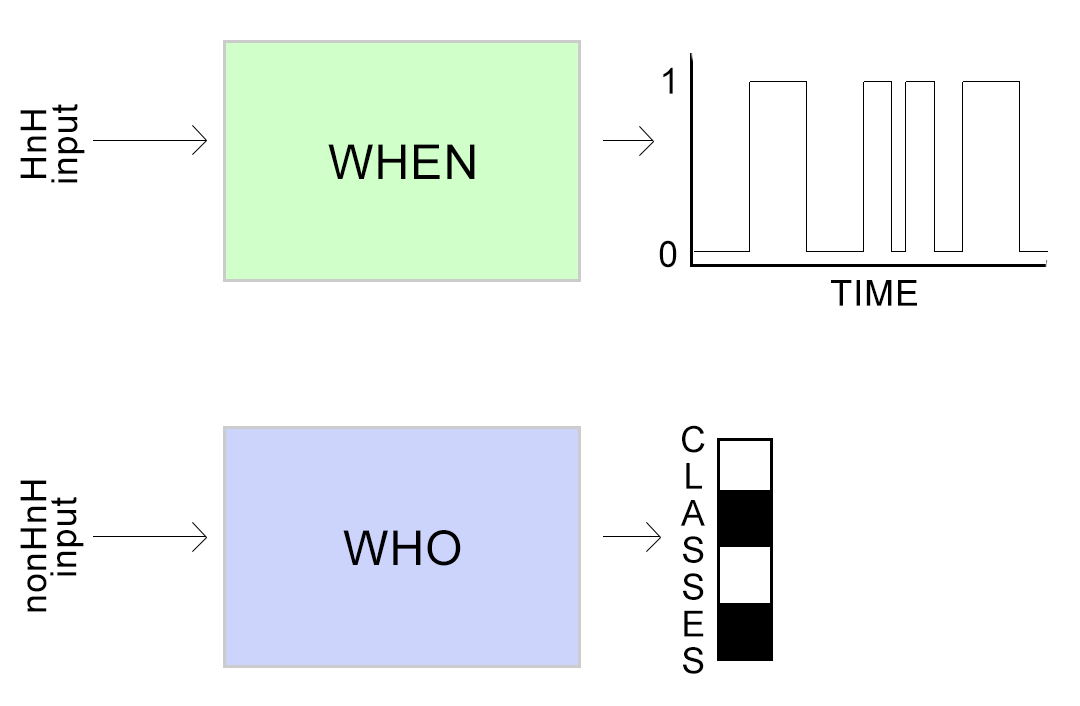}
\caption{Separate training. Networks WHEN and WHO are defined and trained independent of one another, with different types of input.}
\label{fig:sep}
\end{figure}

First, we used separate training for the two tasks. As depicted in Figure \ref{fig:sep}, two independent networks are defined, namely WHEN and WHO with the architectures described in Sections \ref{ssec:when} and \ref{ssec:who} respectively. WHEN network performs audio event detection considering all labels as a single general label, while WHO network performs audio tagging. Different kind of input was used for each network. HnH input was used for WHEN and the normal nonHnH input for WHO. Thus the minibatches used as input for WHO network are randomly generated without taking into account the balance of positive and negative recordings in them. Different types of input is used for each task since they perform differently with different types of input even though the total individual recordings for each one are the same.

The advantage of separate training is that each network can train with the type of input that works better for it. WHEN uses a balanced minibatch of positive and negative recordings (HnH) while WHO uses the conventional random type of minibatch (nonHnH). The main disadvantage of separate training is that each task trains independently of the other. However, these two tasks are somewhat related, hence they should be able to focus the attention of the network to important features and also regularise each other.

\subsection{Joint Training}
\label{ssec:joint}
\begin{figure}[!htb]
\centering
\includegraphics[width=10cm]{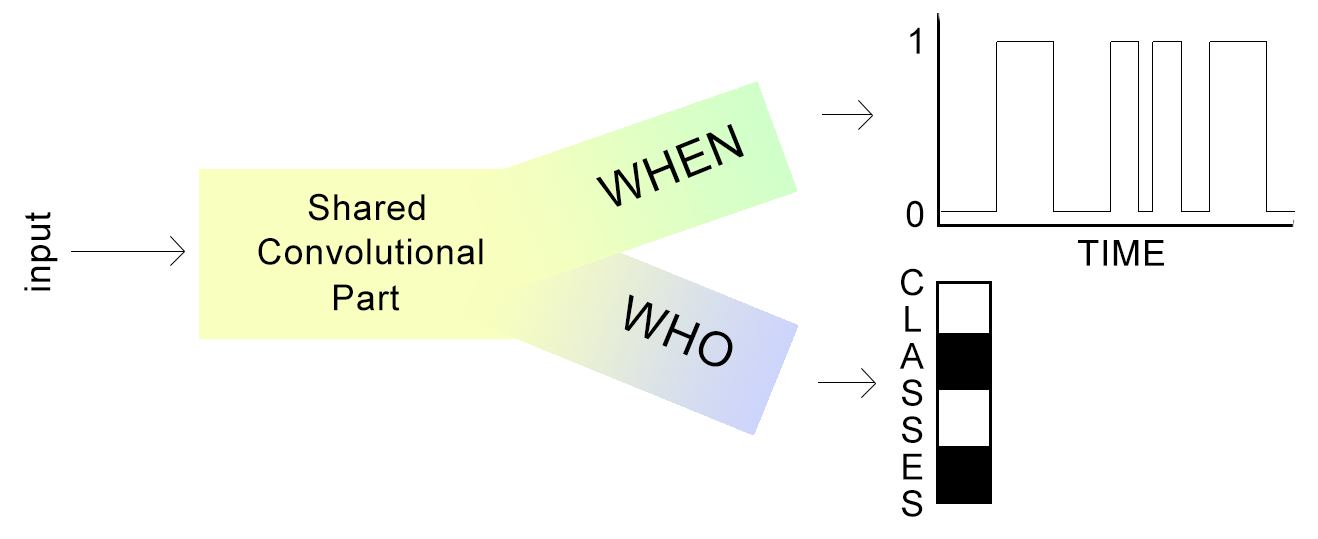}
\caption{Joint training.}
\label{fig:joint}
\end{figure}

Joint training is one of the most common MTL approaches. In joint training the same network is trained for more than one tasks. Usually, the network consists of a few shared layers in the beginning followed by task specific layers before the predictions for each task. For each task a separate loss is computed and then combined into the general loss of the network, usually by weighting each loss. Joint training is a hard parameter sharing approach, since all tasks share the same initial layers and weights. Figure \ref{fig:joint} depicts how our intermediate tasks are adapted to the joint training approach. The \textit{Shared Convolutional Part} consists of the common convolutional and max pooling layers while the separate branches of the network consist of the task specific layers for WHEN and WHO as described in Tables \ref{tab:when} and \ref{tab:who}, respectively. 

The advantages of joint training are all the advantages presented by MTL. More specifically, information is shared between the tasks to help alleviate the issue of limited data. The model focuses its attention on features that are more relevant to all tasks. Also, it reduces the risk of overfitting, since one task can act as the other's regulariser. One of the disadvantages of joint training is that both tasks train on the same input which depending on the type of it (HnH or nonHnH) degrades the performance of one of the tasks (WHO or WHEN respectively), as we will show in Section \ref{sec:eval}.

\subsection{Tied Weights Training}
\label{ssec:tied}
\begin{figure}[!htb]
\centering
\includegraphics[width=10cm]{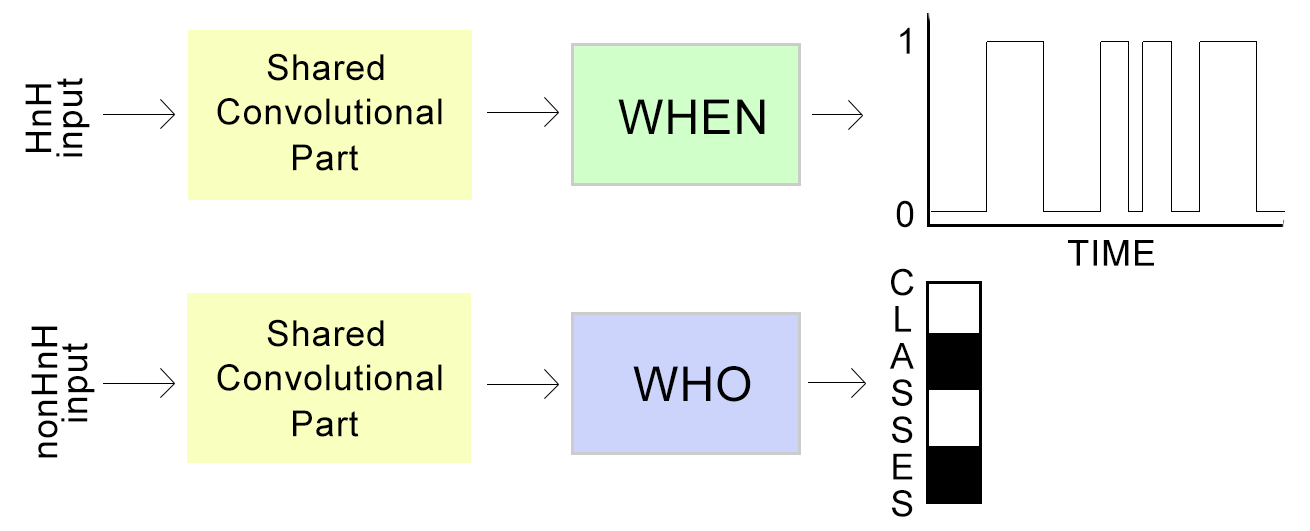}
\caption{Tied weights training.}
\label{fig:tied}
\end{figure}

In order to achieve the advantages of both separate and joint training without any of their disadvantages, we propose a new approach of MTL. Tied weights training follows the hard parameter training convention, where layers and their weights are shared between tasks. However, in contrast to joint training different types of input can be used to train each task. Figure \ref{fig:tied} depicts the structure of tied weights training. \textit{Shared Convolutional Part} refers to the common convolutional and max pooling layers of WHEN and WHO, and shares the same weights between the two tasks. Each network is trained consecutively for one epoch, updating the weights of the shared layers. Using this approach, one can train each network with independent types of input as in separate training while keeping all the advantages of MTL learning.

\section{Evaluation}
\label{sec:eval}
In order to test our approach in a low-resource dataset we use the training dataset provided during the Neural Information Processing Scaled for Bioacoustics (NIPS4B) bird song competition of 2013 that is publicly available and contains 687 recordings of maximum length of 5 seconds each.\footnote{\url{http://sabiod.univ-tln.fr/nips4b/challenge1.html}} For the NIPS4B dataset the recordings have already been weakly labelled and the labels are provided by the organisers along with the dataset recordings. The dataset contains a total of 87 classes, with each being active in only 7 to 20 recordings. Each recording has 0 to 6 classes active in it. Such a dataset can be considered low-resource since the total amount of training time is less than one hour and also there are 87 possible labels that have very sparse activations, 7 to 20 positive recordings for each. 

For our experiments, we split the NIPS4B 2013 training dataset into a training set and testing set. During the NIPS4B 2013 bird song competition, only the weak labels for the training dataset were released, hence we could only use these recordings and couldn't make any use of the NIPS4B 2013 testing dataset that consisted of more recordings. We acquired the strong labels of most of the training dataset recordings via manual annotations, to be used only for evaluation purposes.\footnote{\url{https://figshare.com/articles/Transcriptions_of_NIPS4B_2013_Bird_Challenge_Training_Dataset/6798548}} For our training set the first 499 recordings of the NIPS4B 2013 training dataset are used, while the rest are included in our testing set, excluding 14 recordings for which confident strong annotations could not be attained. Those 14 recordings were added to our training set totalling to 513 training recordings and 174 testing recordings.

In order to efficiently use the data provided by the NIPS4B 2013 training dataset for our WHEN task, we first consider all 87 unique labels as one general label `bird' and train an audio event detection network for this class. Another limitation of this dataset is the imbalance of positive and negative recordings: out of the whole dataset (687 recordings) only 100 of them are labelled as negative (not having any bird present in them). We provide a balanced training set by using our Half and Half training approach. For this dataset, HnH will randomly duplicate the negative recordings during training in order to balance their amount with the positive recordings.

\subsection{Results}
\label{ssec:results}
The same parameters are used for training both WHEN and WHO network for all three different approaches. Our batchsize is equal to 8 recordings. We use the Adam optimiser \cite{Kingma:15} with a learning rate scheduler that reduces the initial rate of 1e-5 by half every 20 epochs until it reaches a minimum rate of 1e-8. The loss function used for the predictions of the WHEN network is the proposed MMM loss, while we use a binary cross-entropy loss for the multi-class predictions of the WHO network.

First, we trained WHEN and WHO independently. WHEN was trained with a HnH input, since not using HnH can cause the network to either ignore negative recordings or mix the negative and positive frames in a recording. On the other hand, WHO was trained with the conventional nonHnH input since using HnH for WHO made its performance worse. This is due to the fact that the active classes are already very sparse (0 to 6 active classes out of 87 per recording) and for the NIPS4B dataset the HnH input duplicates negative recordings, hence decreases the activation rate for each class, making it even harder to predict.

Next, we trained two versions of the joint network, one of them uses a HnH input while the other a nonHnH input. When training the joint network with HnH the WHO predictions tend to not have a satisfactory performance due to the increase in negative recordings. When training the joint network with the nonHnH input the WHEN task performance is degraded. The loss value of the WHO task tends to be an order smaller than the one for WHEN, hence we trained with two different combination of weights for the task. For one of them both task losses have the same weight of 0.5, while for the other one the weight for the WHO task loss in an order larger than the WHEN, more specifically we used weight 0.5 for WHEN loss and 5.0 for WHO loss. 

Finally, we performed a tied weights training. This solved the issue of using only one type of input since it can train with both HnH and nonHnH input separately for each task as if the tasks are trained independently, while still sharing the weights of the shared layers like the joint training. 

Table \ref{tab:eval} shows the area under the ROC curve (AUC) results for each training approach. We can see that even though the tied weights training has a better overall performance compared to the joint training, separate training still has the best overall results. The best overall results for joint training were produced when using weights 0.5 and 5.0 for WHEN and WHO loss, respectively and also using nonHnH input. Hence, we can conclude that the WHO network is sharing important information with the WHEN network that can boost its performance when enough weight is given to its loss. As mentioned before, any type of joint training has so far been proven to outperform independent training which is not the case in our experiments, when comparing results for both WHEN and WHO. We consider the two tasks to be closely related and use hard parameter sharing approaches. However, the tasks might be more loosely related than we originally considered and a soft parameter sharing approach \cite{Duong:15, Misra:16, Yang:17, Ruder:17} may increase performance.

\begin{table}[!htb]
\caption{Area under the ROC curve (AUC) for the predictions of all training approaches. [WHEN: xx; WHO: yy] indicate the weights xx for WHEN task loss and yy for WHO task loss that were used during joint training.}
\centering
\begin{tabular}{lcccc}
\hline
\textbf{Training} & \textbf{Input Type} &  \textbf{WHEN}	 & \textbf{WHO}\\
\textbf{Method} & \textbf{WHEN $\mid$ WHO} &  \textbf{AUC}	 & \textbf{AUC}\\
\hline
Separate &	HnH $\mid$ nonHnH & \textbf{0.90} & \textbf{0.94} \\
Joint [WHEN: 0.5; WHO: 0.5] & HnH & 0.89  & 0.52 \\
Joint [WHEN: 0.5; WHO: 0.5] & nonHnH & 0.47  & 0.57 \\ 
Joint [WHEN: 0.5; WHO: 5.0] & HnH & 0.90  & 0.50 \\
Joint [WHEN: 0.5; WHO: 5.0] & nonHnH & 0.82  & 0.75 \\ 
Tied Weights & HnH $\mid$ nonHnH & 0.87   & 0.77  \\
\hline
\end{tabular}
\label{tab:eval}
\end{table}

\section{Discussion}
\label{sec:discussion}
In this paper, we present a way to factorise the task of full transcription into multiple intermediate tasks in order to improve performance for low-resource datasets. We propose two intermediate tasks of audio event detection on a single class and audio tagging, referred to as WHEN and WHO task respectively. Additionally, we introduce a balanced input training and a new loss function in the multi instance learning (MIL) setting for the WHEN task. We train these tasks with three different approaches. Firstly, an independent training for each task and then two multi-task learning (MTL) approaches that use hard parameter sharing. One of them is the most commonly used joint training and the other one is our proposed tied weights training. In order to evaluate our approaches we trained each network using a low-resource dataset for birdsong transcription. Our results show that even thought our proposed tied weights training outperforms joint training for these tasks, separate training still performs better than both.

For our future plans, we first intend to explore if soft parameter sharing in MTL can further improve the performance of our intermediate tasks. Then we plan to use the intermediate transcription to boost the performance of a full transcription network. To our current knowledge and based on our latest experiments, trying to perform full transcription without any intermediate tasks for this low-resource dataset did not provide any usable results. Hence, we will attempt to achieve a satisfactory performance when using the intermediate transcription to focus the attention of the full transcription network.

\vspace{6pt} 

\bibliography{stage2_bibliography}
\bibliographystyle{abbrv}

\end{document}